%% file: google_main.tex
\documentclass[11pt, a4paper, copyright, goog]{google}

\usepackage[authoryear, sort&compress, round]{natbib}
\input{math_commands.tex}

\usepackage{algorithm}
\usepackage{algorithmic}
\usepackage{array}
\usepackage{arydshln}
\usepackage{wrapfig}
\usepackage[normalem]{ulem}
\usepackage{multirow}
\usepackage{fancyvrb}
\usepackage[suppress]{color-edits}

\newcommand{\sysname}{\textsc{TEAL}}
\newcommand{\qgenname}{\textsc{TEAL-Q}}

\newcommand{\codechip}[3]{{\setlength{\fboxsep}{1.5pt}\colorbox{#1}{\textcolor{#2}{\texttt{\small #3}}}}}
\definecolor{toolgreen}{rgb}{0.05,0.45,0.15}
\definecolor{toolbg}{rgb}{0.92,0.97,0.92}
\newcommand{\tool}[1]{\codechip{toolbg}{toolgreen}{#1}}
\definecolor{opred}{rgb}{0.65,0.05,0.05}
\definecolor{opbg}{rgb}{0.99,0.93,0.93}

\definecolor{cmdblue}{rgb}{0.15,0.25,0.55}
\definecolor{cmdbg}{rgb}{0.93,0.95,0.99}
\newcommand{\cmd}[1]{\codechip{cmdbg}{cmdblue}{#1}}

\definecolor{steporange}{rgb}{0.75,0.35,0.0}
\newcommand{\cmark}{\textcolor{toolgreen}{\ding{51}}}
\newcommand{\xmark}{\textcolor{opred}{\ding{55}}}
\newcommand{\trajstep}[1]{\textcolor{steporange}{\textbf{#1}}}
\newcommand{\trajtool}[1]{\textcolor{toolgreen}{\textbf{#1}}}
\definecolor{thinkpurple}{rgb}{0.45,0.15,0.55}
\definecolor{outgray}{gray}{0.4}
\newcommand{\trajthink}[1]{\textcolor{thinkpurple}{\textit{#1}}}
\newcommand{\trajout}[1]{\textcolor{outgray}{#1}}
\newcommand{\trajlegend}{\noindent{\ttfamily\trajstep{[step]}}~step \quad {\ttfamily\trajtool{tool}}~evolved tool \quad {\ttfamily input}~Worker input \quad {\ttfamily\trajthink{Thought}}~Worker thinking \quad {\ttfamily\trajout{-> output}}~tool or shell output\par\vspace{3pt}}
\definecolor{figgrey}{rgb}{0.639,0.635,0.616}
\definecolor{figblue}{rgb}{0.165,0.471,0.839}
\definecolor{figorange}{rgb}{0.922,0.408,0.204}

\makeatletter
\g@addto@macro\ps@firststyle{%
  \fancyfoot[L]{%
    \footerfont\textsuperscript{\textdagger}Work done while interning at Google.\\
    Emails: \texttt{yyang063@usc.edu}, \texttt{jianizhang@google.com}%
    \ifthenelse{\boolean{copyright}}{\\ \copyrightext}{}%
  }%
}
\makeatother

\title{Unsupervised Tool Evolution for Coding Agents in Data-Lake Analytics}

\uselogo{} 

\author[1,\textdagger]{Yuqing Yang}
\author[2,3]{Yurong Liu}
\author[1,2]{Deqing Fu}
\author[2]{Manav Garg}
\author[2]{Sravan Babu Bodapati}
\author[2]{Fatma Ozcan}
\author[2]{Alon Y. Halevy}
\author[2]{Jiani Zhang}

\affil[1]{University of Southern California}
\affil[2]{\thepa{}}
\affil[3]{New York University}

\begin{abstract}
\input{sections/abstract}
\end{abstract}

\begin{document}

\maketitle

\input{sections/introduction}

\bibliography{iclr2026_conference}

\clearpage
\appendix
\input{sections/appendix}

\end{document}

%% file: math_commands.tex
\usepackage{amsmath,amsfonts,bm}

\def\eqref#1{equation~\ref{#1}}

\def\1{\bm{1}}

\DeclareMathAlphabet{\mathsfit}{\encodingdefault}{\sfdefault}{m}{sl}
\SetMathAlphabet{\mathsfit}{bold}{\encodingdefault}{\sfdefault}{bx}{n}



%% file: sections/abstract.tex
Coding agents can answer analytical questions over a data lake by discovering, cleaning, and joining the relevant data. However, they discard this work after each question and redo it from scratch for the next one. Tool evolution, which automatically builds and refines reusable tools across tasks, is a natural fit for this repetition. Since a ground-truth label for a data-lake question requires an expert to solve it first, we consider the unsupervised setting. We propose \sysname{} (\textbf{T}ool \textbf{E}volution for \textbf{A}nalytics over data \textbf{L}akes), which evolves a \textit{library of tools and usage skills} for a given Worker agent without labels. Two agents propose tools from complementary evidence, an Explorer from the data lake itself and a Critic from Worker trajectories. A proposed change is kept only if the Worker adopts the tool on a rerun without extra steps and any changed answer is supported by trajectory evidence. We also introduce \qgenname{}, which generates synthetic questions for tool evolution when real ones are unavailable. We evaluate on KramaBench and LakeQA with four Worker LLMs under step budgets, which cap the number of steps the Worker may take per question. With a tight budget of 10 or 15 steps, \sysname{} improves every Worker on both benchmarks, lifting Sonnet-5 at 15 steps by 9.3 points on KramaBench and 21.7 points on LakeQA. A tighter budget also yields better feedback for evolution. For example, Sonnet-5's LakeQA library evolved at 20 steps and deployed unchanged at 30 steps scores 68.3, above both the baseline (59.6) and a library evolved directly at 30 steps (64.2). Additionally, a library evolved on synthetic questions performs comparably to one evolved on real questions. These results show that useful tools can be evolved without labels, and hopefully inspire future work on what to evolve and how to obtain feedback in unsupervised agent evolution.

%% file: sections/introduction.tex
\section{Introduction}
\label{sec:intro}

A data lake~\citep{nargesian2019data,hai2023data} stores a large, heterogeneous collection of tables, documents, and raw files. Answering analytical questions over a data lake requires discovering relevant data~\citep{fernandez2018aurum,castelo2021auctus}, transforming messy records into structured representations~\citep{kandel2011wrangler,li2023auto}, and integrating data across sources~\citep{freire2025large,zhang2026agentic}. Coding agents equipped with a bash shell and a file editor~\citep{wang2024codeact,wang2025openhands} can automate this workflow~\citep{narayan2022can,fernandez2023large}. However, they typically treat each question over the same data lake in isolation and discard every parser, schema check, and join script as soon as the question ends.

Consider an analyst asking which metropolitan area accounts for the largest proportion of its state's fraud reports~\citep{lai2026kramabench}. The answer lies in 52 per-state CSV files of metropolitan-area counts with irregular headers, hidden among the data lake's thousands of files next to similar national and state summaries. Without a catalog or index, table granularity cannot be inferred from filenames, so a coding agent may pick a national summary table that lacks state breakdowns. Even when the agent locates the 52 state files, opening and cleaning them one by one exhausts a 30-step budget and fails sooner under a 15-step budget. When the next question comes, the agent repeats the same manual exploration and cleaning from scratch.

Database systems avoid repeated scans of raw data by maintaining catalogs~\citep{halevy2016goods}, indexes~\citep{graefe2024more}, and views~\citep{halevy2001answering}. Coding agents can follow the same principle with reusable \emph{tools} that replace repeated step-by-step exploration with a single call. Some of these tools are backed by precomputed \emph{data artifacts}, such as catalogs and indexes, and \emph{usage skills} tell the agent when and how to invoke each tool.

However, existing methods cannot automatically generate such tools, for two reasons.  
First, catalog and ontology formats~\citep{zhang2026autoddg} store descriptive metadata rather than executable tools for cleaning and querying data, whereas code tool generators~\citep{wang2024trove,li2026yunjue} build tools for the task at hand, not artifacts that pay off only across many questions.
Second, supervised tool generators~\citep{cai2024latm} and self-evolving ontologies~\citep{chong2026evoontology} rely on ground-truth validation labels, which are costly to obtain in data-lake analytics because verifying even one answer requires an expert to manually discover, clean, and join the underlying data sets first. Without ground-truth labels, keeping every tool that runs without a runtime error~\citep{li2026yunjue} is unreliable because a tool can execute cleanly while reading a table at the wrong granularity, dropping rows during parsing, or joining tables on unaligned keys.

To address these challenges, we introduce \sysname{} (\textbf{T}ool \textbf{E}volution for \textbf{A}nalytics over data \textbf{L}akes), an unsupervised framework that evolves a \emph{library} for a fixed Worker agent. The library consists of executable tools, some backed by precomputed data artifacts, and usage skills. At test time, the library is frozen and provided to the Worker to answer new questions.

\subsection{Contributions}
\paragraph{Tool evolution from data exploration and Worker trajectories.}
\sysname{} builds the library offline from two complementary channels, which differ in the evidence they draw on. An \emph{Explorer} agent inspects the data lake without seeing any question or trajectory and builds tools that a domain expert would prepare for the data lake, for example search and lookup tools backed by precomputed catalogs or cleaned tables. A \emph{Critic} agent examines the Worker's trajectories on training questions to find where the Worker wastes steps or makes errors, and creates or revises tools accordingly. A \emph{Skill Writer} agent then updates the usage skills to reflect every implemented change.

\vspace{-10pt}
\paragraph{Tool verification without ground-truth answers.}
Because a newly proposed tool can still contain logic bugs or be ignored by the Worker, we introduce a rerun verification stage that accepts or rejects each change without answer labels. \sysname{} reruns the Worker with the updated library on the questions that motivated the change. The change is kept only if the Worker uses the changed tool, still finishes without extra steps, and, whenever its answer changes, the trajectory evidence supports the new answer over the old one.

\vspace{-10pt}
\paragraph{Synthetic question generation for tool evolution.}
When real user questions are scarce for a data lake, we introduce \qgenname{} to generate questions that can be used by \sysname{}, using a small set of seed questions as demonstrations. \qgenname{} first drafts a question from a cluster of related files, then hardens it over several revisions by adding data processing and reasoning demands, such as multi-table joins, while ensuring it is answerable.

\vspace{-10pt}
\paragraph{Empirical evaluation across models and step budgets.}
We evaluate \sysname{} on KramaBench~\citep{lai2026kramabench} and LakeQA~\citep{wang2026lakeqa} with four Worker backbones (Gemini-3.8-Flash, Gemini-3.1-Pro, Sonnet-5, and Opus-4.8) under varying step budgets, i.e., the maximum Worker steps per question. We make three main observations.
First, under tight 10- and 15-step budgets, \sysname{} improves accuracy for every Worker on both benchmarks. With Sonnet-5 at 15 steps, it gains 9.3 points on KramaBench and 21.7 points on LakeQA, and also outperforms the execution-only tool evolution method~\citep{li2026yunjue}.
Second, libraries evolved under a tighter budget transfer to looser budgets. Sonnet-5's LakeQA library evolved at 20 steps scores 68.3 at 30 steps, above both the baseline (59.6) and a library evolved at 30 steps (64.2), suggesting that tighter budgets give better feedback for evolution.
Third, evolving on 50 synthetic questions from \qgenname{} performs comparably to evolving on real questions, and our unsupervised rerun-based verification matches oracle ground-truth verification. The library evolved in the OpenHands harness also transfers unchanged to Google ADK, Codex CLI, and Prime Agent, improving accuracy by up to 16.3 points.

\section{Related Work}
\label{sec:related-work}

\paragraph{Tools and Agents for Data Analytics.}
Data analytics over heterogeneous sources involves discovering relevant files~\citep{paton2023dataset}, preparing and integrating their contents~\citep{liu2024magneto}, and performing analysis~\citep{nam2025ds,yang2026llms}. Prior work supports this with resources that humans prepare in advance, including data artifacts such as metadata catalogs and relationship indexes~\citep{halevy2016goods,fernandez2018aurum} or materialized views and indexes~\citep{agarawal2000automated}, and tools such as user-defined plugins~\citep{qiao2023taskweaver} or a fixed toolset for searching and inspecting the lake~\citep{wang2026lakeqa}. Recent work lets agents build such resources themselves. In EvoOntology~\citep{chong2026evoontology}, a builder agent constructs and refines an ontology layer that data agents query through tools, but a refinement is kept only if the score on a held-out labeled split improves. Closest to our setting, DataCOPE~\citep{qiu2026unsupervised} discovers skills without labels by scoring sampled trajectories through answer agreement, but its skills are procedural guidance, not callable tools backed by persistent artifacts. We study fully automatic tool evolution, where no human writes tools, curates artifacts, or labels answers, and agents decide from data exploration and unlabeled trajectories which tools to build and how to guide their use with skills.

\vspace{-5pt}
\paragraph{Tool Evolution for LLM Agents.}
Automatic tool creation for LLMs began with math and science questions that a single program can answer. A model writes a helper function, verified against labeled examples~\citep{cai2024latm} or by execution tests~\citep{lu2026tte,shen2026uct}, and reuses it across questions of the same type. Agent systems extended this to tools built during long-horizon tasks~\citep{xia2025liveswe} and usually stored for later use. These agents create a tool when a task cannot proceed without one, such as a transcript downloader or a web page fetcher, so the tool is exercised immediately, and acceptance reduces to whether it executes~\citep{li2026yunjue}, whether the environment reflects completion~\citep{wang2023voyager}, or whether the agent judges its output plausible~\citep{qiu2025alita,ding2026evosop}. In this work, we study unsupervised tool evolution for data-lake analytics, where neither assumption holds. Here tools rarely fill a gap, since the Worker is not blocked without them, and they cannot be credited for a correct answer without labels. \sysname{} therefore verifies tools by the Worker's behavior on a rerun, and we study how to make unsupervised tool evolution more effective, in particular by evolving under tighter step budgets.

\section{Method}
\label{sec:method}

\subsection{Problem Formulation}
\label{sec:setting}

\paragraph{Coding agents and tools.}
We consider coding agents equipped with a shell and a file editor, which enable them to read and modify files and execute code. Beyond these basic capabilities, we define a \emph{tool} as a named, callable program added to the agent's action space and retained across tasks. Once implemented and debugged, a tool can be invoked in a single action without reconstructing its procedure. A tool may also rely on persistent artifacts, such as indexes or cached metadata, that are built once and reused across tasks.

\paragraph{Tool evolution for data-lake analytics.}
Given a natural-language question over a data lake, a coding agent, which we call the \emph{Worker}, autonomously discovers the relevant files and performs the data preparation, integration, analysis, and reasoning needed to answer the question under a step budget. Without access to ground-truth answers, a tool evolution system observes the Worker's trajectories on training questions and produces a \emph{library} $\mathcal{L} = (\mathcal{T}, S)$, consisting of a set of tools $\mathcal{T}$ and a skill file $S$, a natural-language document that tells the Worker when and how to use the tools. We then freeze the library and measure the Worker's accuracy with it on a disjoint set of test questions over the same data lake.

\begin{figure}[t]
    \centering
    \includegraphics[width=\textwidth]{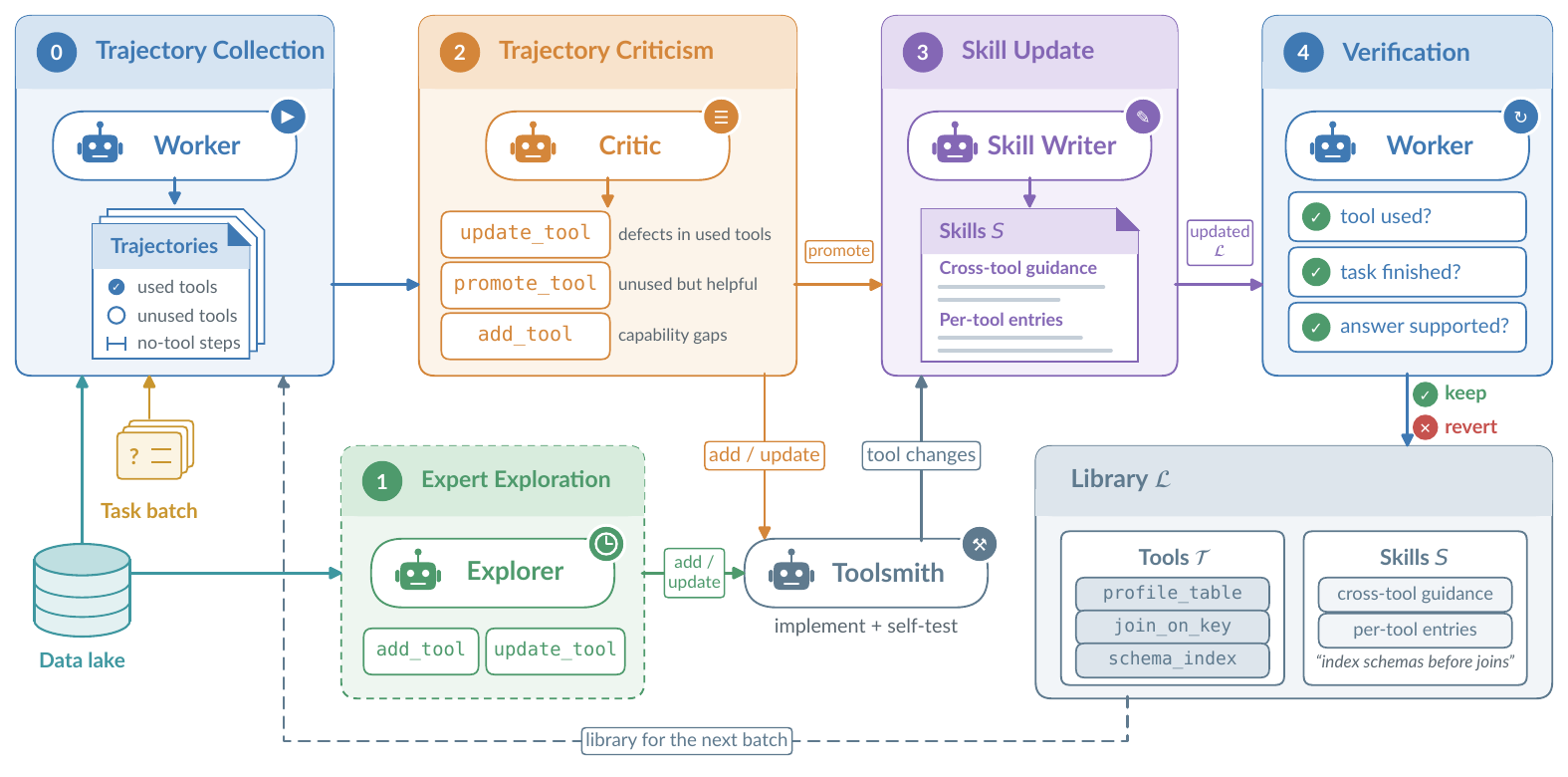}
    \caption{The \sysname{} evolution loop. The Worker first solves the questions with the current library (Stage 0). The Explorer proposes tools for the data lake as a whole (Stage 1), and the Critic proposes targeted fixes from the Worker's trajectories (Stage 2). The Skill Writer then updates the usage guidance (Stage 3), and a rerun of the Worker decides which changes to keep (Stage 4).  Robot icons mark LLM roles. No stage uses ground-truth answers.}
    \label{fig:toolevo_pipeline}
    \vspace{-10px} 
\end{figure}

\subsection{Unsupervised Tool Evolution via \sysname{}}
\label{sec:toolevo}

Alongside the \textbf{\emph{Worker}} that answers questions over the data lake, \sysname{} coordinates five LLM agents to evolve  $\mathcal{L}$. The \textbf{\emph{Explorer}} inspects the data lake to propose data-specific tools and precomputed data artifacts. The \textbf{\emph{Critic}} diagnoses execution bottlenecks and errors in the Worker trajectories to propose targeted tool changes. The \textbf{\emph{Toolsmith}} implements proposals from the Explorer and Critic as executable code and validates them with unit tests. The \textbf{\emph{Skill Writer}} maintains $S$ that guides when and how to invoke each tool in $\mathcal{T}$, and the \textbf{\emph{Analyst}} compares original and rerun trajectories during verification whenever their final answers differ.

\subsubsection{The tool evolution pipeline}
During offline tool evolution, which we refer to as training even though no model weights are updated, \sysname{} processes the training questions in batches and runs the five stages in Figure~\ref{fig:toolevo_pipeline} on each batch to iteratively update $\mathcal{L}$.

In \textbf{Stage 0: Trajectory Collection}, the Worker executes a batch of questions with the current library $\mathcal{L}$. Each trajectory is recorded together with a summary of its tool usage as evidence for later stages. The summary covers which tools were used, which were unused, and where the Worker took many consecutive steps without any tool call. For example, on a question over a large directory or an Excel workbook, the summary flags that the Worker spent seven consecutive bash steps running file searches without calling any library tool and exhausted its step budget.

In \textbf{Stage 1: Expert Exploration}, an Explorer agent inspects the data lake and $\mathcal{L}$ without access to task questions or Worker trajectories. It \emph{acts as a domain expert} preparing the tools for the data lake as a whole rather than for any specific task, and proposes \texttt{add tool} and \texttt{update tool} operations. It can also create tools backed by persistent artifacts that precompute information for reuse across tasks. A Toolsmith agent implements these proposals and checks them with executable unit tests. Unlike the other stages, this stage runs only once every $E$ batches. For example, after surveying thousands of raw CSV and Excel files, the Explorer proposes a dataset search tool backed by a precomputed SQLite catalog of file formats, column headers, and sample rows.

In \textbf{Stage 2: Trajectory Criticism}, a Critic agent reviews the evidence collected in Stage 0 and the current tool set $\mathcal{T}$, including those just added or updated by the Explorer, to propose \texttt{update tool} operations for defects in used tools, \texttt{promote tool} operations for unused tools that could have helped, and \texttt{add tool} operations for capability gaps that the current tools do not cover. The \texttt{add tool} and \texttt{update tool} proposals are implemented by the Toolsmith, and the \texttt{promote tool} proposals are passed to the next skill update stage. For instance, when trajectories show the Worker failing on Excel sheets whose headers do not start on the first row, the Critic proposes \texttt{add tool} for a header-aware workbook loader, or proposes \texttt{update tool} when an existing catalog tool misses yearly editions of a table.

In \textbf{Stage 3: Skill Update}, a Skill Writer agent updates the skill file $S$ by integrating the implemented changes from the Explorer and the Critic and the \texttt{promote tool} proposals. $S$ contains two parts, the overall cross-tool guidance and the per-tool entries. For example, the Skill Writer adds a cross-tool workflow rule instructing the Worker to query the catalog tool before running shell searches, and documents when to invoke the workbook loader on multi-sheet Excel files.

In \textbf{Stage 4: Verification}, the Worker reruns the questions with the updated library. Each rerun is compared with the original run on three checks, whether the changed tool was used, whether the task still finished, and whether any changed answer is better supported by the trajectory (Section~\ref{sec:verify}). A change is kept if a rerun passes all three checks and reverted otherwise, after at most one round of repair. The resulting library $\mathcal{L}$ carries forward to the next batch and is frozen after offline training for evaluation.

\paragraph{Two complementary channels for tool evolution.}
\label{sec:channels}
\sysname{} combines two complementary channels, the Critic and the Explorer, because neither suffices alone. Prior tool-evolution methods rely only on trajectory analysis similar to our Critic~\citep{xia2025liveswe,li2026yunjue,ding2026evosop}. Trajectory analysis is effective for fixing specific execution errors, such as misread offset headers in Excel sheets or bugs in called tools. However, because the Critic only sees a small batch of trajectories at a time, it focuses on immediate fixes for those questions rather than building data lake-wide tools. The Explorer addresses this limitation. Without seeing any question or trajectory, it is instructed to act as a domain expert preparing tools for the whole data lake. It scans the lake's directory structures, file formats, and schemas and proposes tools that serve many questions, such as search and lookup tools, some backed by precomputed data artifacts such as metadata catalogs and value indexes. These tools let the Worker locate relevant data in a single tool call. In short, the Critic asks \emph{what should change given how the Worker actually solved these tasks?}, while the Explorer asks \emph{what would a domain expert build for this data lake?}

\subsubsection{Verification without ground-truth answers}
\label{sec:verify}

A proposed tool change can pass its unit tests and still fail to help the Worker. The tool may contain a logic error, such as reading a table at the wrong granularity, or the Worker may simply ignore it. \emph{Verification} is the stage that decides whether to keep or revert each change. 
Because ground-truth answers are unavailable, \sysname{} cannot judge a change by whether it makes answers correct. Instead, it reruns the Worker with the updated library on the questions that motivated the change and compares each rerun with the original trajectory on three checks.

The first check confirms that the Worker actually calls the changed tool during the rerun, since a tool that is never called cannot improve any answer. When the Worker ignores the tool, the reason is usually unclear guidance rather than a broken tool, so \sysname{} asks the Skill Writer to revise the tool's usage instructions and reruns the question before rejecting the change.

The second check confirms that the change does not make the Worker worse at finishing the question. A rerun fails this check if it produces no answer where the original run produced one, or if it uses more than $m$ steps beyond the $n$ steps of the original run.

The third check applies only when the rerun produces a different answer from the original run. In this case, an Analyst agent compares the two trajectories and judges which answer is better supported by the execution evidence, for example whether the final value appears in a tool output or is merely asserted by the Worker. The Analyst does not assume that the original answer is correct. The rerun fails this check only if the evidence favors the original answer, and passes if the evidence favors the new answer or is inconclusive.

\sysname{} keeps a change if at least one rerun passes all three checks and reverts it otherwise. In this way, verification goes beyond checking whether a tool runs without errors and uses the Worker's own trajectories as evidence of whether the tool actually helped.

\begin{figure}[t]
    \centering
    \includegraphics[width=\textwidth]{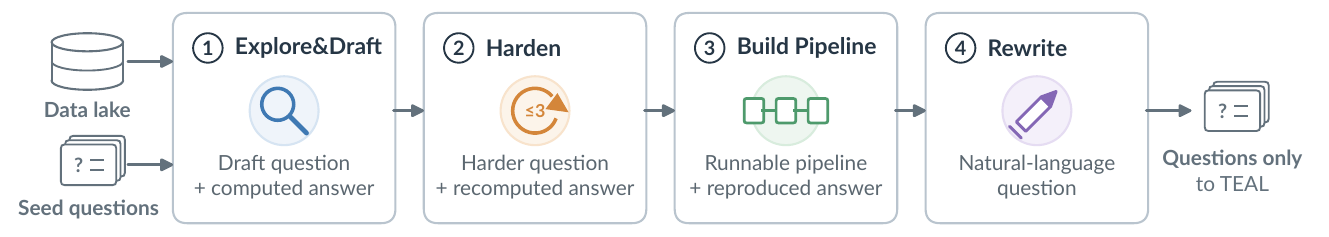}
    \caption{Synthetic question generation with \qgenname{}. Given the data lake and a small set of seed questions as demonstrations, \qgenname{} generates each question in four stages.}
    \label{fig:qgen_pipeline}
    \vspace{-10px} 
\end{figure}

\subsection{Synthetic Question Generation via \qgenname{}}

Questions written by users are scarce and not always accessible. Existing data lake benchmarks contain only around a hundred questions each, and a newly deployed data lake may have none.
Because \sysname{} does not need ground-truth answers, evolution needs only questions. We therefore propose \qgenname{}, which generates synthetic questions from the data lake, using a small set of seed questions as demonstrations of style and difficulty.
As shown in Figure~\ref{fig:qgen_pipeline}, \qgenname{} generates each question in four stages. 

\textbf{\textcircled{1} Explore and Draft.}   \qgenname{} inspects a cluster of relevant data files and drafts a question with an answer, computed by running code on the data. 
\textbf{\textcircled{2} Harden.} In up to three rounds, \qgenname{} makes the question harder, for example by adding a join across tables or a hard-to-parse file, and recomputes the answer after each round.
\textbf{\textcircled{3} Build Pipeline.} \qgenname{} turns the final solution into an executable pipeline and checks that it reproduces the computed answer. This confirms that the question is answerable from the data.
\textbf{\textcircled{4} Rewrite.} \qgenname{}  rewrites the question in natural language, preserving essential definitions that are needed to make it unambiguous but omitting details that the Worker can discover itself.

\qgenname{} passes only the questions to \sysname{}. The answers and solution pipelines serve only to keep each question well posed and answerable. As a result, \qgenname{} does not need to verify that its answers are correct, which is a major challenge in synthetic data generation~\citep{ramrakhya2025autoplay,gureja2025verification}.

\section{Experiments}
\label{sec:experiments}

Our experiments address two research questions (RQs). \textbf{RQ1: Are agent-evolved tools useful?} We ask whether a library evolved without labels improves a coding agent's accuracy on data-lake analytics, across Workers, datasets, and step budgets. \textbf{RQ2: How can feedback be used effectively without ground-truth answers?} We ask which signals, from step budget pressure to the proposal channels and rerun-based verification, make unsupervised evolution work.

\subsection{Experimental Settings}

\paragraph{Datasets.}

For KramaBench~\citep{lai2026kramabench}, we merge the data lakes of all six domains (archaeology, astronomy, biomedical, environment, legal, and wildfire) into a single unified data lake of 1,742 files and use stratified sampling to select 51 training and 53 test tasks.

For LakeQA~\citep{wang2026lakeqa}, we use its tables-only variant, which keeps the 130 questions whose final reasoning step relies on tables and folds any fact from removed non-table sources into the question text, preserving the original answers. We randomly split them into 50 training and 80 test tasks. Each requires one to five reasoning hops over a lake of 4,180 tables.

We report LakeQA accuracy under its deterministic scorer and the official KramaBench score, which is exact match for most tasks and partial credit (F1 for lists, relative error for approximate numerics, an LLM judge for free-text answers) for the rest. We refer to both as accuracy.

\begin{figure*}[t]
    \centering
    \includegraphics[width=\textwidth]{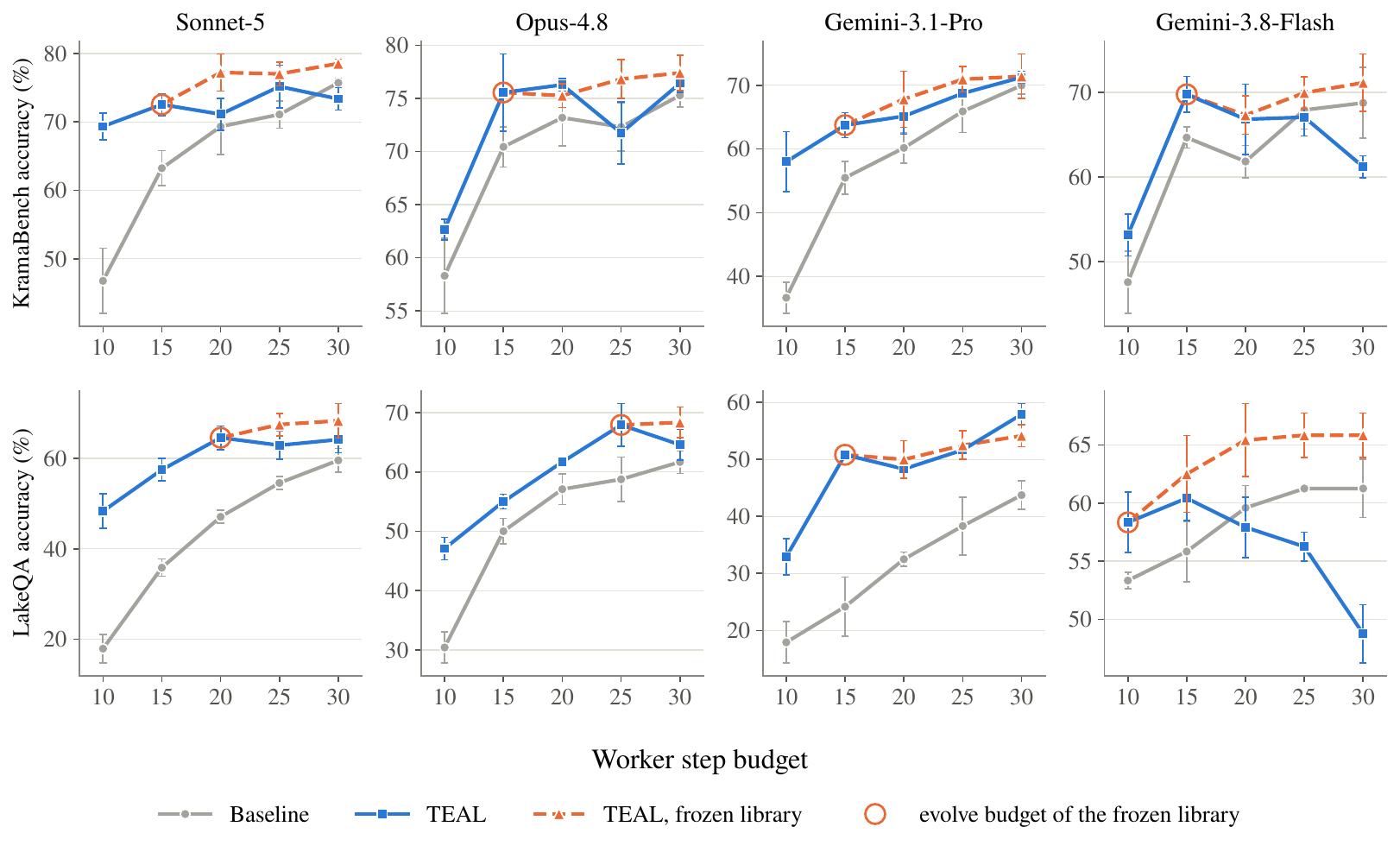}
    \vspace{-18px}
    \caption{Accuracy versus Worker step budget in the real-question scenario. Rows correspond to benchmarks (top: KramaBench, bottom: LakeQA) and columns to Worker models. The horizontal axis is the step budget $k$ at test time, from 10 to 30 steps. The grey line is the baseline Worker without evolved tools. The blue line is \sysname{} with a library evolved and tested at the same budget $k$. The orange dashed line is a frozen library: it is evolved once at the ringed budget and then tested unchanged at each looser budget.}
    \label{fig:step_budget}
    \vspace{-12px} 
\end{figure*}

\paragraph{Setup.}

We use OpenHands\footnote{\url{https://github.com/OpenHands/software-agent-sdk}.} as the harness for all agents. To assess generalization across Worker backends, we evaluate Gemini-3.8-Flash~\citep{google2026gemini38flash}, Gemini-3.1-Pro~\citep{google2026gemini31pro}, Sonnet-5~\citep{anthropic2026sonnet5}, and Opus-4.8~\citep{anthropic2026opus48}, while using Opus-4.8 for all evolution roles. Each Worker episode has a step budget of $k$: up to $k-3$ steps for the task, plus at most 3 steps to write an answer if none has been written by then. A step budget is the practical way to bound the cost and latency of answering a question, and how many steps a deployment can afford varies. We therefore treat it as an experimental variable. In all experiments, we use batches of 5 questions, run the Explorer on the first batch and every $E=4$ batches after, and set $m=\max(2,\,0.2n)$.

\subsection{Main Results}
\label{sec:main-results}

Given the real training questions from the datasets, we sweep $k$ over 10, 15, 20, 25, and 30 steps. Figure~\ref{fig:step_budget} plots accuracy against $k$ for the baseline without the evolved library and for \sysname{}, reporting the mean and standard deviation over three test repeats. We have the following findings.

\textbf{Under a tight step budget, \sysname{} improves accuracy substantially.} With Sonnet-5 as the Worker, \sysname{} improves accuracy by 22.6 points on KramaBench and 30.4 points on LakeQA at a budget of 10 steps, and by 9.3 and 21.7 points at 15 steps. The gain holds for all four Workers on both lakes at 10 and 15 steps. In five of the eight panels, \sysname{} at 15 steps is on par with or above the baseline at 30 steps. We attribute the improvement to the fact that a single tool call replaces a multi-step procedure, such as locating, loading, and joining the relevant files, so the Worker can finish the task within a budget that the baseline exhausts on file discovery. Appendix~\ref{sec:case-study} illustrates this with two test tasks, and Appendix~\ref{sec:tool-stats} lists the most used tools.

\textbf{Under a loose budget, a library evolved at that budget does not always help, but one evolved under a tighter budget does.} At 20 to 30 steps, the library evolved at the deployment budget (blue line) sometimes falls below the baseline (grey line), whereas the library evolved under a tighter budget and deployed unchanged (orange line) stays above the baseline at every looser budget and is usually the best of the three. The difference lies in the feedback that evolution receives. Under a tight budget, the Worker fails in visible ways, through empty answers, truncated pipelines, and repeated processing work, which give the Critic concrete targets. A saved step then changes the rerun outcome, so verification can tell a useful tool from a useless one, and the short trajectories are also easier to analyze. Under a loose budget, the Worker solves most tasks inline, so a tool rarely changes the outcome, and the long trajectories make proposals harder to ground and reruns harder to compare, which leaves the quality of tools and skills unchecked. Step budget pressure thus acts as a training signal in place of gold labels, which makes evolving under a tighter budget and deploying under a looser one a practical recipe.

Appendix~\ref{sec:token-cost} further reports the number of steps each Worker actually uses and the resulting cost per episode, which reflect the practical value of \sysname{} in deployment.

\subsection{Comparison with Existing Methods}
\label{sec:comparison}

We also compare \sysname{} with the agentic methods evaluated in KramaBench and LakeQA and with an existing tool evolution method. \textbf{DS-Guru}~\citep{lai2026kramabench} writes a complete Python script from a data preview and repairs it with error feedback. \textbf{Smolagents DeepResearch}~\citep{roucher2025smolagents, lai2026kramabench} is a ReAct agent~\citep{yao2023react} that executes Python step by step with file inspection tools. \textbf{LakeQA Agent}~\citep{wang2026lakeqa} is the reference agent released with LakeQA, which calls human-designed tools to search, inspect, and run Python, and \textbf{LakeQA Agent + RAG}~\citep{wang2026lakeqa} adds a retrieval tool backed by a pre-built index of the lake's tables. \textbf{ISE}~\citep{li2026yunjue} evolves tools without gold labels and keeps every tool that runs without errors, and we reproduce it from the released code and replace only its Worker with that of \sysname{}.

\begin{wraptable}{r}{0.5\textwidth}
    \vspace{-\intextsep}
    \centering
    \small
    \begin{tabular}{lcc}
    \toprule
         & $k{=}15$ & $k{=}30$ \\
    \midrule
    \multicolumn{3}{l}{\textbf{\textit{KramaBench}}} \\
    DS-Guru & \multicolumn{2}{c}{39.73} \\
    Smolagents & \multicolumn{2}{c}{46.69} \\
    \hdashline[2pt/2pt]\noalign{\vskip 1pt}
    Baseline & $63.23_{\pm 2.57}$ & $75.71_{\pm 0.70}$ \\
    ISE & $66.81_{\pm 5.96}$ & -- \\
    \sysname{} & $72.54_{\pm 1.59}$ & $\textbf{78.53}_{\pm 0.65}{}^{\dagger}$ \\
    \midrule
    \multicolumn{3}{l}{\textbf{\textit{LakeQA}}} \\
    LakeQA Agent & \multicolumn{2}{c}{55.00} \\
    \quad + RAG & \multicolumn{2}{c}{58.75} \\
    \hdashline[2pt/2pt]\noalign{\vskip 1pt}
    Baseline & $35.83_{\pm 1.91}$ & $59.58_{\pm 2.60}$ \\
    ISE & $25.83_{\pm 1.91}$ & -- \\
    \sysname{} & $57.50_{\pm 2.50}$ & $\textbf{68.33}_{\pm 3.82}{}^{\dagger}$ \\
    \bottomrule
    \end{tabular}
    \caption{Accuracy (\%) of existing methods and \sysname{}, Sonnet-5 as the Worker. $^\dagger$Library evolved at 15 (KramaBench) or 20 (LakeQA) steps, deployed unchanged (orange lines in Figure~\ref{fig:step_budget}).}
    \label{tab:main}
    \vspace{-10px} 
\end{wraptable}

For each existing method, we use the hyperparameters from its original evaluation on the corresponding dataset, including a 20-step budget for Smolagents and a 25-step budget for LakeQA Agent, so its result is the same in both columns. Table~\ref{tab:main} leads to two observations.

First, our baseline Worker is already a strong coding agent. On KramaBench, it outperforms DS-Guru and Smolagents by a wide margin even at 15 steps, fewer than the 20 steps Smolagents is allowed. On LakeQA, it surpasses LakeQA Agent at 30 steps, with or without RAG, even though LakeQA Agent comes with human-designed tools and a pre-built retrieval index. \sysname{} improves on this strong baseline on both datasets and at both budgets.

Second, we compare \sysname{} with ISE, another unsupervised tool evolution method whose released code allows a faithful reproduction. ISE retains a tool once it executes successfully. Under the same Worker and the same 15-step budget, ISE underperforms \sysname{} on both datasets (by 5.7 points on KramaBench), indicating that a tool that executes successfully does not necessarily help with the task. This issue is more pronounced on LakeQA, where ISE falls 10.0 points below the baseline. For instance, ISE evolved \tool{search\_file\_content}, which scans the lake line by line. The tool was retained after a single successful run on a simple search, yet 121 of its 126 calls timed out at test time. The steps consumed by these calls prevented the Worker from reaching an answer within the step budget. Under execution-only verification, such a tool therefore does more harm than good.

\subsection{Results in the synthetic-question scenario}
\label{sec:synthetic}

\begin{table}[ht]
    \centering
    \begin{tabular}{lcccc}
    \toprule
         & \multicolumn{2}{c}{KramaBench} & \multicolumn{2}{c}{LakeQA} \\
    \cmidrule(lr){2-3} \cmidrule(lr){4-5}
         & Acc. & $\Delta$ & Acc. & $\Delta$ \\
    \midrule
        Baseline & $63.23_{\pm 2.57}$ & -- & $35.83_{\pm 1.91}$ & -- \\
        \sysname{} (real questions) & $72.54_{\pm 1.59}$ & +9.31 & $57.50_{\pm 2.50}$ & +21.67 \\
        \sysname{} (synthetic questions) & $72.18_{\pm 3.22}$ & +8.95 & $56.25_{\pm 3.31}$ & +20.42 \\
    \bottomrule
    \end{tabular}
    \caption{Accuracy (\%) of \sysname{} evolved on real or synthetic questions with Sonnet-5 as the Worker. All configurations use a 15-step budget and the same held-out test sets.}
    \label{tab:qgen}
    \vspace{-5px}
\end{table}

In the synthetic-question scenario, \qgenname{} generates 50 questions from the lake of each dataset. Evolution sees only these questions, without answers, and the resulting library is evaluated on the same held-out test tasks as in the real-question scenario. In Table~\ref{tab:qgen}, the library evolved on synthetic questions improves Sonnet-5 over the baseline on both datasets and performs comparably to the library evolved on real questions. The same holds for all four Workers (Table~\ref{tab:qgen-full} in Appendix~\ref{sec:qgen-full}). Although the generated questions are never verified by a human, they exercise enough of the lake's preparation and discovery work for \sysname{} to build a library that transfers to real questions, so tool evolution can begin before any user question has been asked.

\subsection{Ablation and Generalization}
\label{sec:ablation}

\begin{wraptable}{r}{0.53\textwidth}
    \vspace{-\intextsep}
    \centering
    \small
    \begin{tabular}{lcc}
    \toprule
         & \multicolumn{2}{c}{KramaBench} \\
    \cmidrule(lr){2-3}
         & Acc. & $\Delta$ \\
    \midrule
        Baseline & $63.23_{\pm 2.57}$ & -- \\
        \sysname{} & $72.54_{\pm 1.59}$ & +9.31 \\
    \midrule
        \quad $-$ Explorer & $71.59_{\pm 2.28}$ & +8.36 \\
        \quad $-$ Evolved skills & $68.69_{\pm 0.85}$ & +5.46 \\
    \hdashline[2pt/2pt]\noalign{\vskip 1pt}
        \quad $+$ Ground-truth answers & $72.73_{\pm 3.07}$ & +9.50 \\
        \quad $+$ Ground-truth solutions & $75.55_{\pm 3.04}$ & +12.32 \\
    \midrule
        Google ADK: Baseline & $53.95_{\pm 4.41}$ & -- \\
        Google ADK: \sysname{} & $70.20_{\pm 3.42}$ & +16.25 \\
        Prime Agent: Baseline & $64.85_{\pm 1.11}$ & -- \\
        Prime Agent: \sysname{} & $67.43_{\pm 2.41}$ & +2.58 \\
        Codex CLI: Baseline & $66.22_{\pm 3.34}$ & -- \\
        Codex CLI: \sysname{} & $70.57_{\pm 0.96}$ & +4.35 \\
    \bottomrule
    \end{tabular}
    \caption{ Accuracy (\%) of ablations and reference configurations with Sonnet-5 as the Worker at a 15-step budget, in the real-question scenario.}
    \label{tab:ablation}
    \vspace{-10px} 
\end{wraptable}

Table~\ref{tab:ablation} isolates the contribution of each component with Sonnet-5 as the Worker at a budget of 15 steps. Removing the Explorer leaves only the Critic channel, which still improves over the baseline but falls slightly short of the full \sysname{}. Evolution without the Explorer also no longer produces the artifact-backed tools shown in Appendix~\ref{sec:case-study}, consistent with the complementary roles of the two channels (Section~\ref{sec:channels}). Removing the evolved skills causes a larger drop, because the Worker no longer knows how to use the tools. A good tool alone therefore does not guarantee a gain, and the tools and their usage skills must be evolved together. 

We also report two oracle variants that give \sysname{} ground-truth labels during evolution, which our unsupervised setting does not have. Given only the ground-truth answers, \sysname{} performs comparably to its label-free version, so our verification checks already obtain the feedback that the answers would provide. Given the full ground-truth solutions as well, the system knows how to solve the tasks and creates more targeted tools, such as \tool{merge\_entity\_tables}, which joins tables on both state and metropolitan area and matches area names that differ only in punctuation, whereas the label-free library only standardizes state names with \tool{normalize\_join\_state\_keys}.

To test generalization across harnesses, we take the library evolved under OpenHands and deploy it unchanged under three other harnesses, Google ADK\footnote{\url{https://github.com/google/adk-python}.}, Prime Agent\footnote{\url{https://github.com/PrimeIntellect-ai/prime-agent}.}, and Codex CLI\footnote{\url{https://github.com/openai/codex}.}, with Sonnet-5 as the Worker in each. The library improves the Worker over its baseline under every harness, so the evolved tools and skills generalize beyond the harness they were evolved in.

\section{Conclusion}
\label{sec:conclusion}

In this paper, we proposed \sysname{}, a framework that evolves tools and usage skills for a coding agent over a data lake without ground-truth labels, and \qgenname{}, which generates synthetic questions for evolution. \sysname{} proposes tools from both the data lake and the Worker's trajectories, and verifies each change by checking the Worker's tool use and trajectory evidence on a rerun, which performs comparably to verification with ground-truth answers. Our results show that unsupervised tool evolution is most effective under a tight step budget, which reveals where the Worker needs tools. Libraries evolved in this way improve all four Workers under tight budgets and remain effective under looser budgets, and libraries evolved on synthetic questions perform comparably to those evolved on real ones.

\vspace{-10px}
\paragraph{Limitations and future work.} The evolved library of tools and skills introduces additional overhead in two forms. First, it lengthens the prompt that the Worker reads at every step, which increases the input tokens. Second, the Worker tends to reason more at each step, plausibly about which tools to call (Appendix~\ref{sec:token-cost}). Future work may reduce the input overhead by improving cache reuse of repeated prompt content, and the reasoning overhead by internalizing the skills into the Worker.






%% file: sections/appendix.tex
\section{Qualitative Analysis}
\label{sec:case-study}

\paragraph{Tools with build artifacts.} As introduced in Section~\ref{sec:method}, the Explorer may create a tool backed by a persistent artifact that is constructed once, stored alongside the tool, and consulted at call time. We illustrate this mechanism with \tool{search\_dataset\_catalog}, the discovery tool of the Sonnet-5 LakeQA library evolved under a budget of 10 steps.

Its artifact is a catalog of the lake, built in a single pass over every directory that infers the true format of each file from its bytes, records the header, an estimated row count, per-column samples, and data-quality flags, and embeds the name of each directory. The pass takes approximately four minutes of wall-clock time and roughly forty batched embedding calls, and the resulting 93 MB SQLite catalog is stored beside the tool and never rebuilt inside a task. The tool itself is a ranked search over this catalog. It takes a natural-language description of the data sought and returns the best-matching directories. The tool was revised fourteen times, each revision traceable to a Worker trajectory. For instance, after a Worker had to fall back to \cmd{ls | grep} to collect the yearly editions of a table, the tool was revised to return all editions of a matched table together.

Across all experiments, most artifacts are catalogs, which answer which file is relevant, or value indexes, which answer which file contains a given value. What exactly is built varies from one evolution run to another. For example, one value index inverts the cell values themselves and reaches 6 GB, whereas another is a full-text index over names, headers, and sampled cells. Other artifacts also occur, such as manifests that record which files belong together and in what order, and decoded copies of archives that were mis-stored as text.

\paragraph{How tools help.} First, tools save steps, which is why \sysname{} is particularly effective under a tight step budget. At test time the Worker invoked \tool{search\_dataset\_catalog} in all 80 held-out tasks. On the task requesting the 2018 violent crime rate of Albany County, every \sysname{} repeat finished within four steps. One catalog query located the county-level crime-rate table, one SQL call through the companion \tool{query\_tabular\_file} tool returned 359.8, and the Worker submitted the answer. Without such a tool, the baseline used nine to ten steps in every repeat, most of them on \cmd{find} and \cmd{grep} over the lake. The additional steps did not translate into accuracy, since in two of the three repeats the search selected an incorrect crime table and the baseline did not answer correctly. Figure~\ref{fig:albany_traces} shows one repeat of each.

Second, tools make accuracy more robust, because a tool's output is consistent across repeats while the Worker's manual work is not. On the task asking for the calcium reading at the smallest dust-proxy value in a paleoclimate workbook, \tool{inspect\_excel\_workbook} detected the header on the sixth row and returned the columns under their real names, so every \sysname{} repeat filtered the two relevant columns with short scripts and answered correctly. The baseline, with nine steps in every repeat but without reusable tools, did not reach a correct answer. In one repeat it parsed the workbook manually and misaligned the header, and in another it located the file but failed to parse the header. Figure~\ref{fig:workbook_traces} shows one repeat of each.

\section{Tool Statistics}
\label{sec:tool-stats}

Tables~\ref{tab:tools_kb} and~\ref{tab:tools_lakeqa} list the five most used tools in the libraries evolved for each lake with Gemini-3.1-Pro as the Worker. We run each test task three times and call each run an episode. The coverage of a tool is the percentage of test episodes in which the Worker called that tool at least once. On both lakes the most used tools answer where the data lives.

As shown in Table~\ref{tab:tools_kb}, three of the KramaBench tools help the Worker find data. One maps a topic to the files that hold it, one searches for a column name or a value across all files, including Excel workbooks that \cmd{grep} cannot read, and one lists the files in the lake. The other two load files that standard readers parse incorrectly, namely CSVs that stack several tables in one file and Excel sheets whose header is not on the first row. The LakeQA lake is larger, with over four thousand files. As shown in Table~\ref{tab:tools_lakeqa}, its library therefore indexes the lake ahead of time. One tool looks up a pre-built catalog that records where each dataset is and which columns and format it has, and another searches a pre-built full-text index of dataset names and column headers. Its compute tool runs a filter or an aggregate over a large file through DuckDB, which streams the file instead of loading it into memory. The Worker calls the catalog tool in nearly every LakeQA episode, so almost every solution starts by querying the catalog rather than walking the directory tree.

\begin{table}[t]
    \centering
    \footnotesize
    \setlength{\tabcolsep}{4pt}
    \begin{tabular}{@{}lr>{\raggedright\arraybackslash}p{2.3cm}>{\raggedright\arraybackslash}p{4.6cm}@{}}
    \toprule
    Tool & Coverage & Role & What it does \\
    \midrule
    \tool{resolve\_concept\_to\_datasets} & 55.3\% & concept lookup & Maps a natural-language concept or a few keywords to the lake files that hold the corresponding measurements. \\
    \tool{locate\_field\_in\_lake} & 39.6\% & field search & Finds every file, sheet, and column in which a query string appears as a header or a cell value, including inside Excel workbooks that \cmd{grep} cannot read. \\
    \tool{catalog\_data\_lake} & 37.7\% & enumeration & Lists the files under a lake root, following symlinks and optionally filtered by substring, glob, or regex, with a one-line schema per file. \\
    \tool{load\_report\_csv} & 36.5\% & CSV loading & Loads one or more report-style CSVs in one call, detecting stacked data sections separated by title rows and blank rows. \\
    \tool{inspect\_excel\_workbook} & 13.8\% & workbook loading & Lists every sheet of a workbook with its shape and detected header row, or loads one sheet as a clean typed table. \\
    \bottomrule
    \end{tabular}
    \caption{The five most used tools in the KramaBench library evolved with Gemini-3.1-Pro as the Worker under a 10-step budget. Coverage (\%) is the share of the 159 test episodes (53 tasks, three repeats) that call the tool at least once.}
    \label{tab:tools_kb}
\end{table}

\begin{table}[t]
    \centering
    \footnotesize
    \setlength{\tabcolsep}{4pt}
    \begin{tabular}{@{}lr>{\raggedright\arraybackslash}p{2.3cm}>{\raggedright\arraybackslash}p{4.6cm}@{}}
    \toprule
    Tool & Coverage & Role & What it does \\
    \midrule
    \tool{query\_lake\_catalog} & 97.1\% & metadata lookup & Answers where a dataset is, which columns and format it has, how big it is, what its known traps are, or which sibling files it has, from a pre-built SQLite catalog of the lake. \\
    \tool{aggregate\_over\_files} & 82.1\% & streaming compute & Runs one filter, count, distinct, projection, group-by, or aggregate over a large delimited file or a union of sibling files through DuckDB, without loading the file into memory. \\
    \tool{preview\_candidate\_columns} & 40.4\% & schema preview & Shows the columns of a set of candidate files given as paths, a glob, or a dataset slug. \\
    \tool{resolve\_filter\_values} & 32.9\% & value matching & Maps a fuzzy category term to the exact distinct values stored in a column, with the row count of each value. \\
    \tool{search\_lake\_content} & 30.8\% & full-text search & Finds which dataset, file, or column contains a keyword, phrase, or value through a pre-built full-text index over dataset names and column headers. \\
    \bottomrule
    \end{tabular}
    \caption{The five most used tools in the LakeQA library evolved with Gemini-3.1-Pro as the Worker under a 20-step budget. Coverage (\%) is the share of the 240 test episodes (80 tasks, three repeats) that call the tool at least once.}
    \label{tab:tools_lakeqa}
\end{table}

\section{Steps and Cost}
\label{sec:token-cost}

\subsection{Step counting}
\begin{figure*}[ht]
    \centering
    \includegraphics[width=\textwidth]{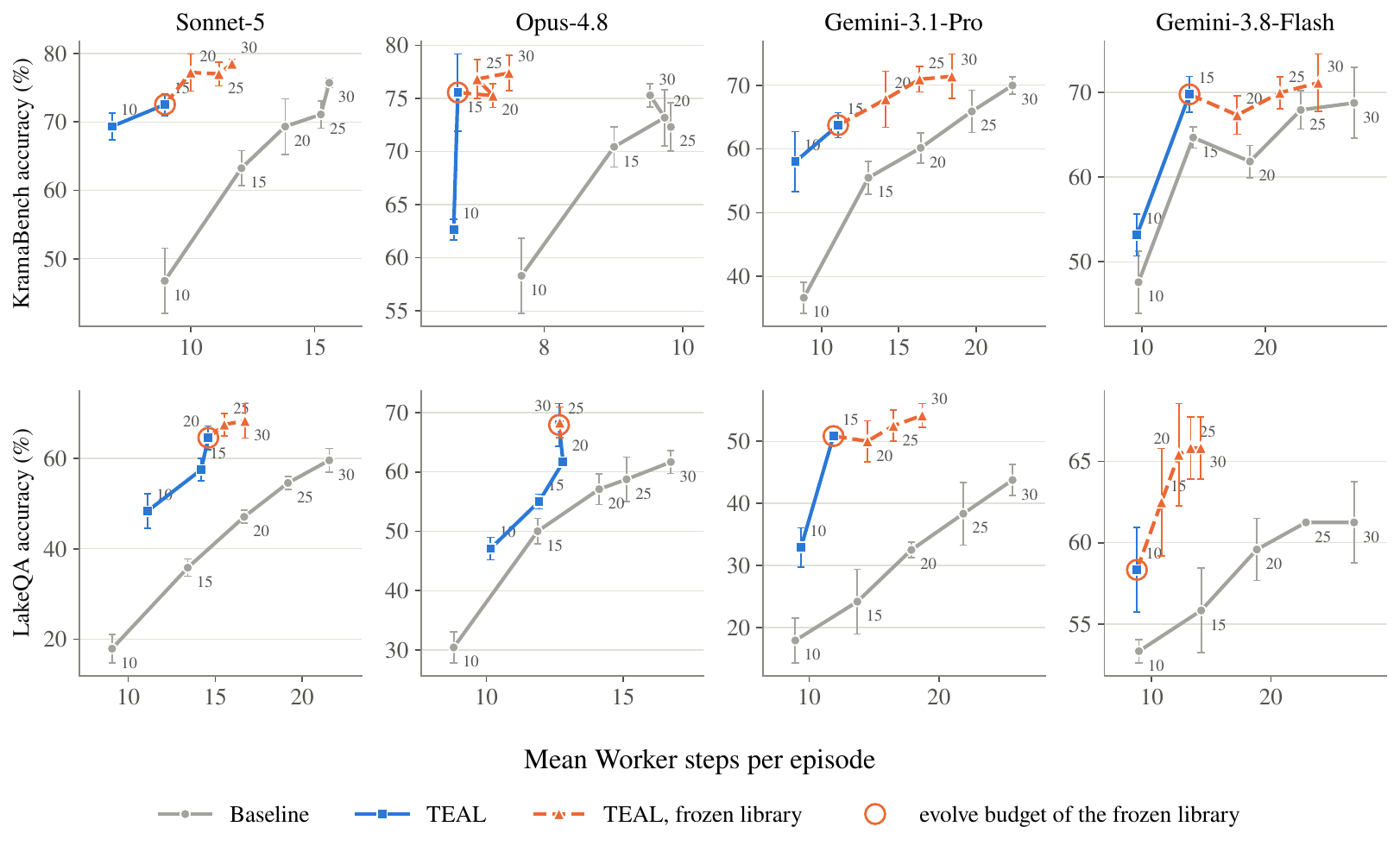}
    \vspace{-18px}
    \caption{Accuracy versus mean Worker steps per test episode.}
    \label{fig:steps_accuracy}
\end{figure*}

The number of steps a Worker actually uses can differ from its step budget. Figure~\ref{fig:steps_accuracy} therefore plots the accuracy of \sysname{} and the baseline against the steps they actually use under each step budget. The horizontal axis is the mean number of steps per episode over three test repeats, and the vertical axis is the accuracy, so a curve closer to the upper left offers a better trade-off between accuracy and steps. As expected, the evolved tools let the Worker reach higher accuracy with fewer steps.

\subsection{Cost accounting}

\begin{figure*}[t]
    \centering
    \includegraphics[width=\textwidth]{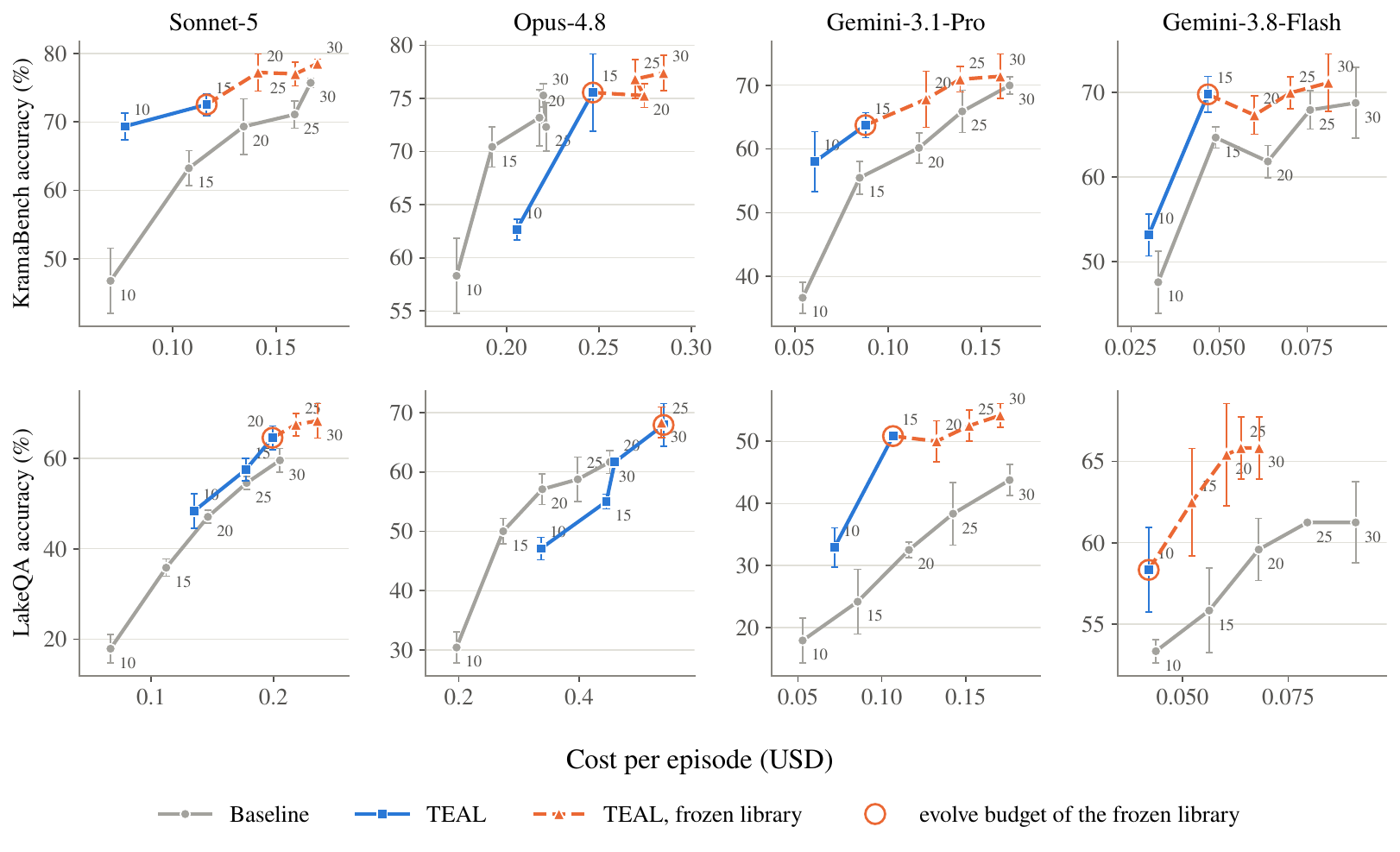}
    \vspace{-18px}
    \caption{Accuracy versus estimated inference cost per test episode.}
    \label{fig:cost_accuracy}
\end{figure*}

A step budget is a practical way to bound the cost and latency of an agent in deployment. A budget stated directly in tokens or in dollars is less practical, since the agent would spend extra tokens tracking how many it has already used. However, a step budget does not reflect the actual cost of an episode. We therefore also plot accuracy against the estimated cost per test episode in Figure~\ref{fig:cost_accuracy}.

\begin{table}[ht]
    \centering
    \begin{tabular}{lccc}
    \toprule
    Worker & Uncached input & Cached input & Output \\
    \midrule
    Sonnet-5 & 2.50 & 0.20 & 10.00 \\
    Opus-4.8 & 6.25 & 0.50 & 25.00 \\
    Gemini-3.1-Pro & 2.00 & 0.20 & 12.00 \\
    Gemini-3.8-Flash & 0.75 & 0.075 & 3.75 \\
    \bottomrule
    \end{tabular}
    \caption{Vertex AI list prices (USD per million tokens) used to estimate the cost of a test episode.}
    \label{tab:prices}
\end{table}

We compute the cost of each test episode from three token counts. The \emph{uncached input} is the number of tokens written to the prompt cache for the first time, the \emph{cached input} is the number of tokens read back from the cache on later turns, and the \emph{output} is the number of completion tokens, including hidden reasoning, which both providers bill as output. We assume an optimal caching policy. The prefix, consisting of the system prompt, the tool schemas, and the skills, stays the same across all test episodes with the same method, Worker, lake, and step budget, so it is always read from the cache. Each turn then writes only the tokens added since the previous turn and reads the rest from the cache. The cost of an episode is then
\begin{equation}
    C = n_{\mathrm{u}}\, p_{\mathrm{u}} + n_{\mathrm{c}}\, p_{\mathrm{c}} + n_{\mathrm{o}}\, p_{\mathrm{o}},
\end{equation}
where $n_{\mathrm{u}}$, $n_{\mathrm{c}}$, and $n_{\mathrm{o}}$ are the uncached input, cached input, and output token counts of the episode, and $p_{\mathrm{u}}$, $p_{\mathrm{c}}$, and $p_{\mathrm{o}}$ are the corresponding Vertex AI list prices of the Worker, given in Table~\ref{tab:prices}. Since it assumes no cache misses, this estimate is a lower bound. We report this estimate instead of the billed cost because prompt caching behaved differently across Workers in our runs, so the billed cost mostly reflects the caching configuration of the client, whereas the estimate applies the same accounting to every Worker and method.

Figure~\ref{fig:cost_accuracy} is read in the same way as Figure~\ref{fig:steps_accuracy}, except that the horizontal axis is the mean cost per episode, so a curve closer to the upper left again offers a better trade-off. With Sonnet-5, Gemini-3.1-Pro, and Gemini-3.8-Flash as the Worker, \sysname{} generally reaches higher accuracy at the same cost, or the same accuracy at a lower cost. Opus-4.8 has the most expensive input tokens, which limits the cost advantage of \sysname{}, but \sysname{} remains the better choice when the goal is higher accuracy.

Three factors shape the cost of \sysname{}. First, \sysname{} adds several tools and their skills to the prompt, so its fixed prefix is roughly two to three times as long as the baseline's, and the Worker reads this prefix again at every step. Prompt caching absorbs much of this overhead, so we recommend enabling it when deploying \sysname{}, and we hope that future work on more effective caching will further reduce the cost of such repeated input. Second, perhaps surprisingly, when \sysname{} shortens an episode, the saving appears in uncached input rather than in output. The result of a tool or shell command is billed as new input at the next step, and an evolved tool returns a compact result in place of the raw intermediate output that the baseline passes through the shell. On KramaBench at 15 steps, for example, \sysname{} cuts the uncached input per episode by 37\% to 63\% for Sonnet-5, Gemini-3.1-Pro, and Gemini-3.8-Flash. Third, the output shrinks much less than the number of steps, since the output per step grows for most Workers. We attribute this to the Worker reasoning more at each step, plausibly about which of the additional tools to call.

\section{Synthetic-Question Scenario Across Workers}
\label{sec:qgen-full}

Table~\ref{tab:qgen-full} extends Table~\ref{tab:qgen} to all four Workers.

\begin{table}[ht]
    \centering
    \small
    \setlength{\tabcolsep}{5pt}
    \renewcommand{\arraystretch}{1.08}
    \begin{tabular}{ll cccc}
        \toprule
        \multirow{2}{*}{Model}
        & \multirow{2}{*}{Methods}
        & \multicolumn{2}{c}{KramaBench}
        & \multicolumn{2}{c}{LakeQA} \\
        \cmidrule(lr){3-4} \cmidrule(lr){5-6}
        & & Acc. & $\Delta$ & Acc. & $\Delta$ \\
        \midrule

        \multirow{3}{*}{Sonnet-5}
        & Baseline
        & $63.23_{\pm 2.57}$ & --
        & $35.83_{\pm 1.91}$ & -- \\
        & \sysname{} (real questions)
        & $72.54_{\pm 1.59}$ & $+9.31$
        & $57.50_{\pm 2.50}$ & $+21.67$ \\
        & \sysname{} (synthetic questions)
        & $72.18_{\pm 3.22}$ & $+8.95$
        & $56.25_{\pm 3.31}$ & $+20.42$ \\
        \midrule

        \multirow{3}{*}{Opus-4.8}
        & Baseline
        & $70.44_{\pm 1.90}$ & --
        & $50.00_{\pm 2.17}$ & -- \\
        & \sysname{} (real questions)
        & $75.55_{\pm 3.62}$ & $+5.11$
        & $55.00_{\pm 1.25}$ & $+5.00$ \\
        & \sysname{} (synthetic questions)
        & $75.73_{\pm 1.26}$ & $+5.29$
        & $55.42_{\pm 1.91}$ & $+5.42$ \\
        \midrule

        \multirow{3}{*}{Gemini-3.1-Pro}
        & Baseline
        & $55.48_{\pm 2.54}$ & --
        & $24.17_{\pm 5.20}$ & -- \\
        & \sysname{} (real questions)
        & $63.74_{\pm 1.97}$ & $+8.26$
        & $50.83_{\pm 0.72}$ & $+26.67$ \\
        & \sysname{} (synthetic questions)
        & $64.22_{\pm 4.86}$ & $+8.74$
        & $47.92_{\pm 4.39}$ & $+23.75$ \\
        \midrule

        \multirow{3}{*}{Gemini-3.8-Flash}
        & Baseline
        & $64.65_{\pm 1.25}$ & --
        & $55.83_{\pm 2.60}$ & -- \\
        & \sysname{} (real questions)
        & $69.75_{\pm 2.12}$ & $+5.10$
        & $60.42_{\pm 1.91}$ & $+4.58$ \\
        & \sysname{} (synthetic questions)
        & $70.50_{\pm 2.80}$ & $+5.85$
        & $59.17_{\pm 1.91}$ & $+3.34$ \\
        \bottomrule
    \end{tabular}
    \caption{Accuracy (\%) of \sysname{} evolved on real or synthetic questions,
    compared with the same Worker without evolved tools.
    All configurations use a 15-step budget and the same held-out test tasks.}
    \label{tab:qgen-full}
    \vspace{-5px}
\end{table}

\begin{figure}[t]
\scriptsize
\fvset{commandchars=@?!, fontsize=\fontsize{7.3}{8.1}\selectfont}
\trajlegend
\noindent\textbf{\sysname{} \cmark}
\begin{Verbatim}
@trajstep?[step 1]! @trajtool?search_dataset_catalog!
  query: "violent crime rate by county New York 2018"
  @trajout?-> {"best_match":!
        @trajout?"index-violent-property-and-firearm-rates-by-county-beginning-1990",!
      @trajout?"dataset_geography_sample": [{"column": "County", "value": "Albany"}, ...],!
      @trajout?"score": 10.104, ...}!
@trajstep?[step 2]! @trajtool?query_tabular_file!
  @trajthink?Thought: This is exactly the dataset. Buffalo (1832) is earlier than!
  @trajthink?Rochester (1834), so Buffalo is the terminus city; the other terminus is!
  @trajthink?Albany, in Albany County.!
  path: "data/datagov/index-violent-property-and-firearm-rates-by-county-beginning-1990/
         files/rows.txt"
  mode: "query"
  sql:  "SELECT County, Year, "Violent Rate" FROM {src}
         WHERE County='Albany' AND Year=2018"
  @trajout?-> {"rows": [{"County": "Albany", "Year": 2018, "Violent Rate": 359.8}],!
      @trajout?"columns": ["County", "Year", "Violent Rate"], "n_returned": 1, ...}!
@trajstep?[step 3]! file_editor  create ./answer.txt  "359.8"
\end{Verbatim}
\vspace{2pt}\hrule\vspace{4pt}
\noindent\textbf{Baseline \xmark, where it wasted steps on data discovery.}
\begin{Verbatim}
@trajstep?[step 1]! $ find ./data -iname "*crime*"; find ./data -iname "*albany*"
  @trajout?-> (no output)!
@trajstep?[step 2]! $ find ./data -type f | head -100
  @trajout?-> (no output)!
@trajstep?[step 3]! $ pwd; ls -la; find . -maxdepth 3 -type d; find . -type f | wc -l; ...
  @trajout?-> lrwxrwxrwx  data -> /.../lake/lakeqa_tables/unified!
@trajstep?[step 4]! $ find -L ./data -type f | wc -l; find -L ./data -iname "*crime*"; \
           find -L ./data -iname "*albany*"
  @trajout?-> 4180!
     @trajout?./data/datagov/crime-incidents-in-2010!
     @trajout?[... 20 more datasets ...]!
     @trajout?./data/datagov/violent-crime-property-crime-by-county-1975-to-present!
     @trajout?[... 16 more datasets ...]!
     @trajout?./data/datagov/city-of-albany-five-year-capital-plan-beginning-2013!
@trajstep?[step 5]! $ find -L \
           ./data/datagov/violent-crime-property-crime-by-county-1975-to-present -type f
  @trajout?-> ./data/datagov/violent-crime-property-crime-by-county-1975-to-present/!
     @trajout?files/rows.txt!
@trajstep?[step 6]! $ head -1 ".../rows.txt"; \
           grep -i "^\"Albany\"\|,Albany," ".../rows.txt" | grep ",2018,"
  @trajout?-> JURISDICTION,YEAR,POPULATION,MURDER,RAPE,ROBBERY,AGG. ASSAULT,B & E,!
     @trajout?LARCENY THEFT,M/V THEFT,GRAND TOTAL,PERCENT CHANGE,VIOLENT CRIME TOTAL,[...]!
     @trajout?(grep matched nothing, exit code 1)!
@trajstep?[step 7]! $ grep -i "Albany" ".../rows.txt" | head -20
  @trajout?-> (no output)!
@trajstep?[warning]! @trajout?All 7 task steps are used. Please write your answer now.!
@trajstep?[step 8]! file_editor  create ./answer.txt  "421.8"
\end{Verbatim}
\caption{Trajectories on the Albany County task from Section~\ref{sec:case-study}, both with Sonnet-5 as the Worker under a 10-step budget. Tool outputs and shell observations are abridged, with elisions marked by \texttt{[...]}. The final \texttt{finish} step is omitted.}
\label{fig:albany_traces}
\end{figure}

\begin{figure}[t]
\scriptsize
\fvset{commandchars=@?!, fontsize=\fontsize{7.3}{8.1}\selectfont}
\trajlegend
\noindent\textbf{\sysname{} \cmark}
\begin{Verbatim}
@trajstep?[step 1]! @trajtool?resolve_concept_to_datasets!
  concept: "dust proxy OR wet-dry index OR calcium"
  @trajout?-> {"candidates": [{"path": "./data/climateMeasurements.xlsx", "relevance": 0.4,!
      @trajout?"kind": "data_table", "ext": "xlsx", "recommended_loader": "inspect_excel_workbook",!
      @trajout?"why": {"content_evidence": {"sheet": "Sheet1", "column_index": 0, "column_name":!
      @trajout?"Data Supplement to: \"A 3 million year index for North African humidity/aridity!
      @trajout?[...]\"", ...}}}, [... 7 more candidates ...]], "n_files_walked": 1742, ...}!
@trajstep?[step 2]! @trajtool?inspect_excel_workbook!
  path: "./data/climateMeasurements.xlsx"   sheet: "Sheet1"
  @trajout?-> {"artifact_path": ".../inspect_excel_workbook/cache/climateMeasurements__Sheet1.parquet",!
      @trajout?"artifact_format": "parquet", "artifact_rows": 8360, "artifact_cols": 30,!
      @trajout?"headerless": false, "header_index": 5, "header_depth": 1, "score": 1.0,!
      @trajout?"columns": ["Site", "Hole", "Core", "Section", "Interval_cm", "Depth below seafloor_m",!
      @trajout?"Splice depth_m", "Age_ky", "Al", "Si", "S", "K", "Ca", "Ti", "Mn", "Fe", "Rb", "Sr",!
      @trajout?"Zr", "Ba", "col_20", "Age_ky.1", "PC1", "PC2", "col_24", "Age_ky.2",!
      @trajout?"ODP 967 Dust proxy", "col_27", "Age_ky.3", "ODP 967 wet-dry index"],!
      @trajout?"dtypes": {"Site": "str", "Hole": "str", "Core": "float64", ..., "Ca": "float64", ...},!
      @trajout?"preview": [...]}!
@trajstep?[step 3]! @trajtool?run_python_script!
    df = pd.read_parquet(".../climateMeasurements__Sheet1.parquet")
    min_dust = df['ODP 967 Dust proxy'].min()
    subset = df[df['ODP 967 Dust proxy'] == min_dust]
    min_wet_dry = subset['ODP 967 wet-dry index'].min()
    final_subset = subset[subset['ODP 967 wet-dry index'] == min_wet_dry]
    print(final_subset['Ca'].iloc[0])
  @trajout?-> 207878.877358207!
@trajstep?[step 4]! @trajtool?run_python_script!
    sorted_df = df.sort_values(by=['ODP 967 Dust proxy', 'ODP 967 wet-dry index'])
    print(sorted_df[['ODP 967 Dust proxy', 'ODP 967 wet-dry index', 'Ca']].head(10))
  @trajout?->       ODP 967 Dust proxy  ODP 967 wet-dry index             Ca!
     @trajout?5370               0.001               0.059866  207878.877358!
     @trajout?5369               0.001               0.396778  197141.431580!
     @trajout?[...]!
@trajstep?[step 5]! $ echo "207878.8774" > ./answer.txt
\end{Verbatim}
\vspace{2pt}\hrule\vspace{4pt}
\noindent\textbf{Baseline \xmark, where it failed to parse the header.}
\begin{Verbatim}
@trajstep?[step 1]! $ find ./data -type f
  @trajout?-> (no output)!
@trajstep?[step 2]! $ ls -la
  @trajout?-> lrwxrwxrwx  data -> /.../lake/kb_all/unified!
@trajstep?[step 3]! $ grep -rilE "calcium|dust proxy|wet-dry" ./data
  @trajout?-> (no output, exit code 1)!
@trajstep?[step 4]! $ grep -RiL "calcium" ./data || grep -Ril "calcium" ./data
  @trajout?-> ./data/noaa_wildfires_variabledescrip.csv!
     @trajout?./data/environmental-justice-populations.csv!
     @trajout?[... 393 more files, climateMeasurements.xlsx among them ...]!
@trajstep?[step 5]! $ cat << 'EOF' > search.py
           import os
           import pandas as pd
           for root, _, files in os.walk('./data'):
               for f in files:
                   if f.endswith('.csv'):
                       df = pd.read_csv(path, nrows=5)  [...]
                   elif f.endswith('.xlsx'):
                       df = pd.read_excel(path, sheet_name=sheet, nrows=5)
                       cols = [str(c).lower() for c in df.columns]
                       if any('calcium' in c for c in cols) or any('dust' in c ...):
                           print("Found in:", path, "Sheet:", sheet)  [...]
           EOF
           python3 search.py
  @trajout?-> [An error occurred during execution.] Cannot execute multiple commands at once.!
@trajstep?[step 6]! $ cat << 'EOF' > search.py && python3 search.py
           [... same script, os.walk('./data', followlinks=True) ...]
  @trajout?-> (no "Found in:" line, exit code 0)!
@trajstep?[step 7]! $ grep -ril "dust proxy" ./data
  @trajout?-> (no output, exit code 1)!
@trajstep?[warning]! @trajout?All 7 task steps are used. Please write your answer now.!
@trajstep?[step 8]! $ echo '0.0000' > ./answer.txt
\end{Verbatim}
\caption{Trajectories on the paleoclimate workbook task from Section~\ref{sec:case-study}, both with Gemini-3.1-Pro as the Worker under a 10-step budget. Tool outputs and shell observations are abridged, with elisions marked by \texttt{[...]}. The final \texttt{finish} step is omitted.}
\label{fig:workbook_traces}
\end{figure}